\documentclass[letterpaper, 10pt, conference]{ieeeconf}  
\IEEEoverridecommandlockouts
\makeatletter
\let\NAT@parse\undefined
\makeatother
\usepackage{graphics} 
\usepackage{graphicx}
\usepackage{grffile} 
\usepackage{amsmath} 
\usepackage{amssymb}  
\usepackage{color}
\usepackage{mathtools} 
\usepackage{booktabs,ragged2e}
\usepackage{breqn}
\usepackage{textgreek}
\usepackage{arydshln}
\usepackage[flushleft]{threeparttable}
\usepackage{tabularx,booktabs}
\usepackage{graphics}
\usepackage{amsmath}
\usepackage[normalem]{ulem} 
\usepackage{pifont}
\usepackage{booktabs}
\usepackage{multirow}
\usepackage{lineno}
\usepackage{cite}
\usepackage{bm}
\usepackage{url}
\usepackage{balance}
\usepackage{xcolor}
\usepackage{makecell}%
\usepackage{algpseudocode}
\usepackage{algorithm}
\usepackage{svg}
\usepackage{soul}

\usepackage{makecell}
\usepackage{multirow}
\let\labelindent\relax
\usepackage{enumitem}
\usepackage[table,dvipsnames]{xcolor}
\usepackage[pagebackref=false,breaklinks=true,colorlinks,bookmarks=false]{hyperref}
\usepackage{comment}
\usepackage{todonotes}
\usepackage{svg}
\hypersetup{
linkcolor=BrickRed
,citecolor=Green
,filecolor=Mulberry
,urlcolor=NavyBlue
,menucolor=BrickRed
,runcolor=Mulberry
,linkbordercolor=BrickRed
,citebordercolor=Green
,filebordercolor=Mulberry
,urlbordercolor=NavyBlue
,menubordercolor=BrickRed
,runbordercolor=Mulberry
}

\usepackage[normalem]{ulem} 
\def\methodname{AVO}

\title{\LARGE \bf
AVO: Amortized Value Optimization for Contact Mode Switching in Multi-Finger Manipulation
}

\author{\small 
Adam Hung\textsuperscript{1}, 
Fan Yang\textsuperscript{1*}, 
Abhinav Kumar\textsuperscript{1*}, 
Sergio Aguilera Marinovic$^{2}$,
Soshi Iba$^{2}$,
Rana Soltani Zarrin$^{2}$,
Dmitry Berenson\textsuperscript{1}%
\thanks{$^{*}$Equal contribution. \textsuperscript{1} Department of Robotics, University of Michigan, Ann Arbor, USA. {\tt\footnotesize \{adamhung, fanyangr, abhin, dmitryb\}@umich.edu}, $^{2}$ Honda Research Institute USA. This work was sponsored by Honda Research Institute USA.}
}

\begin{document}
\maketitle
\thispagestyle{empty}
\pagestyle{empty}
\renewcommand{\baselinestretch}{0.979}

\begin{abstract}


Dexterous manipulation tasks often require switching between different contact modes, such as rolling, sliding, sticking, or non-contact contact modes. 
When formulating dexterous manipulation tasks as a trajectory optimization problem, a common approach is to decompose these tasks into sub-tasks for each contact mode, which are each solved independently.
Optimizing each sub-task independently can limit performance, as optimizing contact points, contact forces, or other variables without information about future sub-tasks can place the system in a state from which it is challenging to make progress on subsequent sub-tasks.
Further, optimizing these sub-tasks is very computationally expensive.
To address these challenges, we propose Amortized Value Optimization (AVO), which introduces a learned value function that predicts the total future task performance. 
By incorporating this value function into the cost of the trajectory optimization at each planning step, the value function gradients guide the optimizer toward states that minimize the cost in future sub-tasks. 
This effectively bridges separately optimized sub-tasks, and accelerates the optimization by reducing the amount of online computation needed.
We validate \methodname\ on a screwdriver grasping and turning task in both simulation and real world experiments, and show improved performance even with 50\% less computational budget compared to trajectory optimization without the value function.

\end{abstract}
\vspace{-.15cm}
\section{Introduction}
\vspace{-.2cm}

Previous work has used trajectory optimization to perform dexterous manipulation tasks such as turning a screwdriver~\cite{yang2024multi, tang2024robotic} and in-hand object reorientation~\cite{pang2023global, jiang2024contact}.
However, applying these methods to long-horizon tasks with a high-dimensional robot system remains a challenge, which generally requires extensive computation for solving the optimization and is prone to converging to poor local minima.    

One approach to simplifying the optimization is to decompose the task into sub-tasks separated by contact mode changes. 
In this paradigm, existing methods must optimize trajectories for each sub-task independently, without considering the likelihood of success in future sub-tasks ~\cite{ kumar2024diffusion}, or rely on complex computation to pass gradient information through contact modes~\cite{pardo2017hybrid, chen2024sequential, aceituno2020global, budhiraja2018differential} which can be computationally expensive.
Independent optimization can result in trajectories that minimize each sub-task cost function, but don't perform as well over the entire task.
For example, a grasp might minimize a grasping cost function, but set the robot up in a way that is not conducive to stable object reorientation.

In addition, the high-dimensional nature of the system and the complex constraints used when optimizing trajectories result in high runtimes for these methods.
While learning-based methods~\cite{chen2022system, qi2023hand, handa2023dextreme} reduce runtime by simply querying a model to compute actions, these models are not constraint-aware and can fail in complex, contact-rich tasks \cite{kumar2024diffusion}.

We aim to address the following questions: (1) How can we ensure that, by the end of one sub-task, both the robot and the object are in a stable state conducive to success in the subsequent sub-task?
(2) How can we use past task execution experience to amortize, or decrease the computation time of, trajectory optimization?
\begin{figure}[t]
    \centering
    \includegraphics[scale=.21]{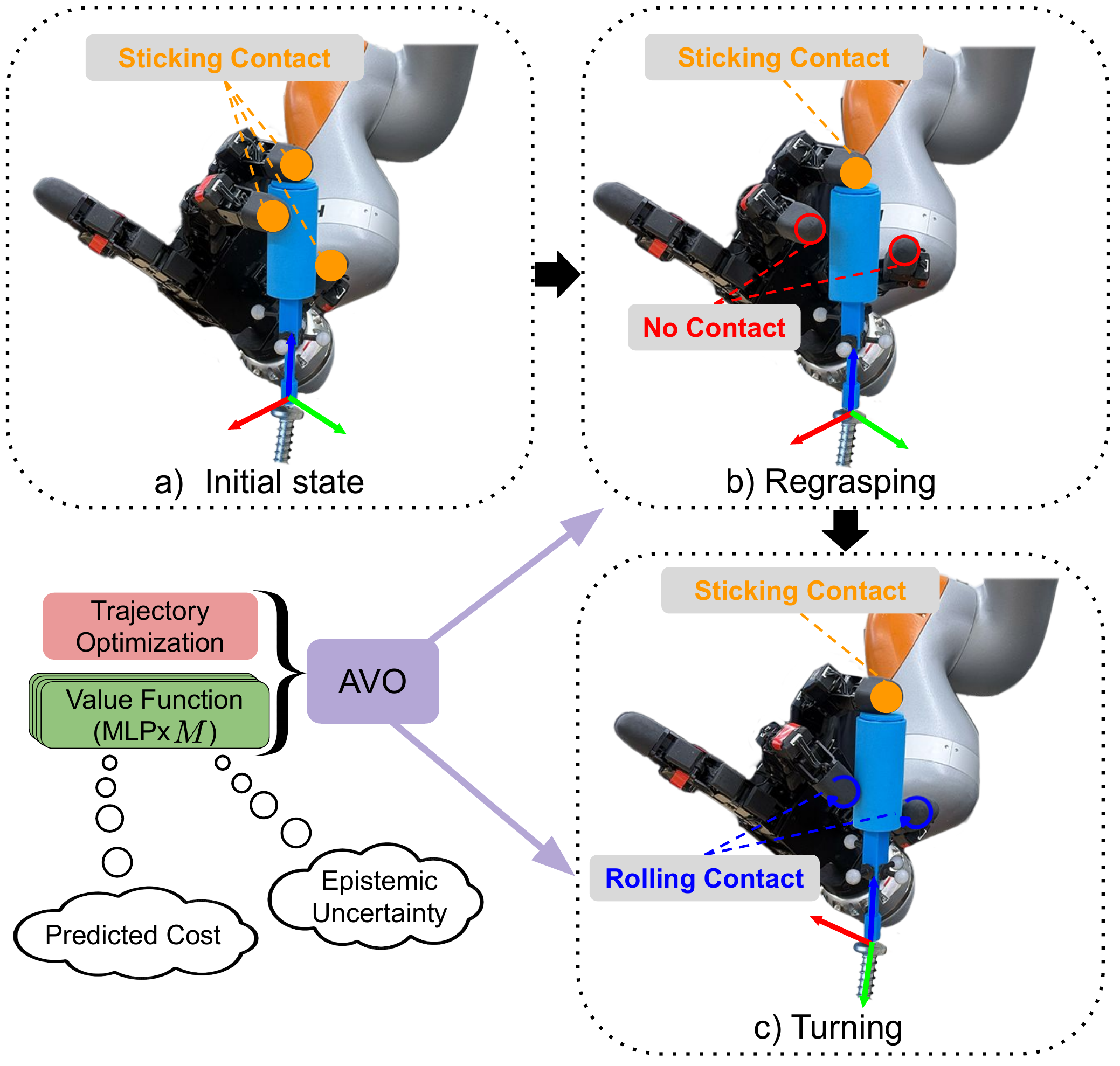}
    \vspace{-.2cm}
    \caption{\methodname\ overview. We train a value function ensemble to predict the mean and variance of the long-horizon future cost of the task. We then use this value function to guide trajectory optimization for a delicate dexterous manipulation task with multiple contact modes. This both improves the quality of the trajectories and also amortizes the optimization.}
    \label{fig:overview}
\end{figure}
To this end, we propose introducing a value function, modeled as an ensemble of neural networks, to use as an auxiliary cost in the trajectory optimization.
Amortized Value Optimization (\methodname) uses a value function to evaluate the quality of every state, providing guidance to both accelerate the optimization and improve the performance of optimized trajectories. 

We show that \methodname\ outperforms baselines that reason independently about individual sub-tasks and reduces the computation time needed to optimize trajectories on challenging screwdriver manipulation tasks in simulation and on hardware.
We also show positive results evaluating against a state-of-the-art amortized learning baseline.

\section{Related Work}

\subsection{Trajectory Optimization with Multiple Contact Modes}
Existing research adopts various approaches to optimize robot trajectories involving multiple contact modes.
Existing research adopts various approaches to optimize robot trajectories that involve multiple contact modes. Contact-implicit trajectory optimization methods~\cite{manchester2020variational, posa2014direct, pang2023global, jin2024complementarity, chen2021trajectotree} jointly optimize both trajectories and contact modes. However, it is usually challenging to solve those contact-implicit problems for a high-dimensional and long-horizon system. To simplify the problem, a common practice is to decompose the tasks into predefined contact modes. 
For example, Stouraitis et al.~\cite{stouraitis2020multi} use compliance modeling to handle the contact mode transition when halting an object. Wang and Kroemer~\cite{wang2019learning} propose a recovery module to ensure robust transitions between desired contact modes. In contrast, our method uses a value function which is trained offline. Chen et al.~\cite{chen2024sequential} adopts a feasibility function to progressively finetune different sub-policies for better long-horizon performances. The main difference is that they use the feasibility function as an auxiliary reward for reinforcement learning, which doesn't enforce constraint satisfaction.

\vspace{-0.2 cm}
\subsection{Trajectory Optimization with Value Functions}
Value functions and Q-functions have been widely used in trajectory optimization and motion planning. One of the key motivations for incorporating value functions is to facilitate long-horizon reasoning, usually by adding the predicted value at the final state to the cost function ~\cite{karnchanachari2020practical, hoeller2020deep, lowreyplan, bhardwaj2020information, sikchi2022learning, hansen2022temporal, parag2022value}. Additionally, Zhong et al.~\cite{zhong2013value} solve a value function-informed, linearly-solvable Markov Decision Process and use the resulting solution to warm start the trajectory optimization. Viereck et al.~\cite{viereck2022valuenetqp} propose a method that uses predicted value function gradients to solve a quadratic program.
Despite these advances, limited research has explored how to use the value function to guide trajectory optimization at every time step of the optimized trajectory. Luck et al.~\cite{luck2019improved} considers each time step, but they do not explore the amortization perspective of using the value function. 
\vspace{-0.1 cm}
\subsection{Amortized Trajectory Optimization}
Amortized trajectory optimization aims to use external information to speed up optimization convergence. A common approach is to train a model to generate a a high-quality trajectory initialization as a warm start~\cite{lembono2020memory, banerjee2020learning, kumar2024diffusion, ichnowski2020deep, natarajan2021learning, celestini2024transformer}. Our method is different in that we train a value function based on an offline dataset to better guide the convergence of the online optimization instead of warm-starting. 

\subsection{Deep Learning for Dexterous Manipulation}
Recent progress has shown the effectiveness of using Reinforcement Learning~\cite{handa2023dextreme, chen2022system, qi2023hand, khandate2024dexterous} and Imitation Learning~\cite{guzey2024bridging, li2024okami, chen2024object} for dexterous manipulation. Unlike our method, these approaches usually do not have explicit reasoning about success-critical constraints, e.g., holding the object without dropping it. 
As a result, they often require non-trivial tuning efforts, without which they exhibit limited robustness in real-world dexterous tasks.

\section{Problem Statement}

We consider the problem of a multi-finger robot hand manipulating an object from initial object configuration $\mathbf{o}_0$ to goal configuration $\mathbf{o}_g$.
The larger manipulation task is split into sub-tasks, where each sub-task has a constant contact mode $c$.
For a multi-finger hand, we can represent the contact mode as a binary vector $\mathbf{c}:=\{0, 1\}^{n_f}$, where $1$ indicates the finger remains in contact throughout the sub-task, and $0$ indicates that contact is broken with the manipulated object before being re-established. 
$n_f$ is the number of fingers.
We refer to fingers in mode $0$ as ``regrasping''. 
Sequencing different contact modes is helpful in contact-rich tasks like turning a screwdriver; in our case, regrasping can readjust the contacts with the screwdriver to enable further turning.

We assume a contact mode sequence $C$ is provided by an upstream planning method such as \cite{kumar2024diffusion}.
Each sub-task has horizon $H$, and the total number of time steps for the full task is $T$.
For a given contact mode, we formulate a trajectory optimization problem with horizon $H \leq T$ as follows,
\vspace{-.2cm}
\begin{align}
\begin{split}
    \mathbf{s}^*_{1:H}, \mathbf{u}^*_{1:H} &= \arg \min J(\mathbf{s}_{1:H}, \mathbf{u}_{1:H}, \mathbf{c}) \\
    &\text{s.t.} \quad   
    h(\mathbf{s}_{1:H}, \mathbf{u}_{1:H}, \mathbf{c}) = 0 \\
    &g(\mathbf{s}_{1:H}, \mathbf{u}_{1:H}, \mathbf{c}) \leq 0,
\end{split}
\label{ps:trajopt_segment}
\end{align}
with state $\mathbf{s}_t$, action $\mathbf{u}_t$,
cost function $J$, and constraint functions $h, g$.
For shorthand, we denote the trajectory $\bm{\tau}:= \{\mathbf{s}_{1:H}, \mathbf{u}_{1:H}\}$. 
To make this problem tractable, we assume (1) access to the geometries of the hand and the object, (2) that the system is quasi-static, and (3) that the fingers begin the task in contact with the object.

\begin{figure*}[t]
    \centering
    \includegraphics[width=\linewidth]{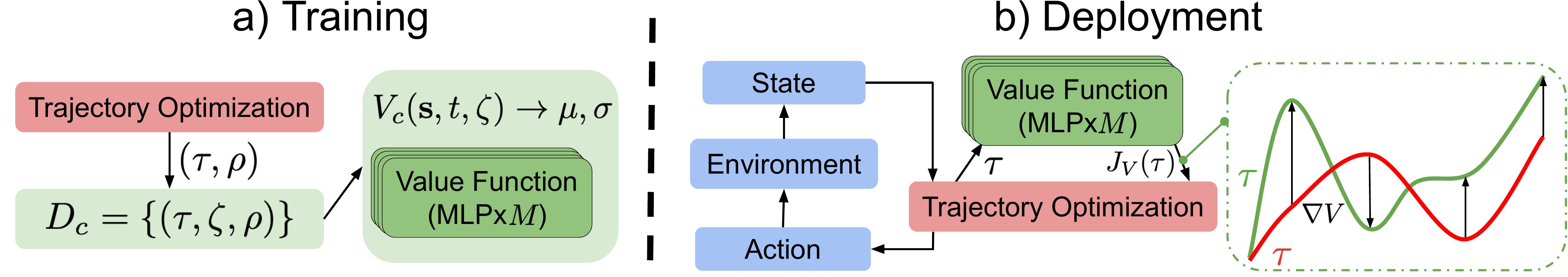}
    \vspace{-.3cm}
    \caption{Training and deployment processes for \methodname. Trajectories $\tau$ for each sub-task with contact mode $c$ are saved, along with the shared final cost. For each contact mode, an ensemble of $M$ value functions are then trained, and at deployment time, the mean $\mu$ and variance $\sigma$ of the ensemble contribute to the optimization cost function.}
    \label{fig:method}
    \vspace{-.7cm}
\end{figure*}

\section{Methods}
\vspace{-.2cm}

In this section, we describe our trajectory optimization and explain how we can train and use a value function to assist in solving the trajectory optimization problem.

\subsection{Trajectory Optimization}

Our trajectory optimization problem is modeled off of \cite{kumar2024diffusion} and \cite{yang2024multi}.
The state $\mathbf{s}$ 
comprises finger configurations $\{\mathbf{q}_i\}_{i=1}^{n_f}$ and object pose $\mathbf{o}$.
The control vector $\mathbf{u}$ is $\{\{\Delta \mathbf{q}_i, \mathbf{f}_i\}^{n_f}_{i=1}, \mathbf{f}_e\}$, where $\mathbf{f}_i$ represents the contact force for the $i$th finger, and $\mathbf{f}_e$ is the force from the environment, with all forces written in the object frame. 
The constraints used in the trajectory optimization problem depend on the contact mode $\mathbf{c} \in \{0, 1\}^{n_f}$, reflecting the specific contact behaviors for each finger.
We partition the state and control vectors into contact fingers $\{\mathbf{s}_c, \mathbf{u}_c\} = \{\mathbf{q}_i, \Delta \mathbf{q}_i, \mathbf{f}_i : \mathbf{c}_i = 1\}$, regrasping fingers $\{\mathbf{s}_r, \mathbf{u}_r\} = \{\mathbf{q}_i, \Delta \mathbf{q}_i : \mathbf{c}_i = 0\}$, and the object and environment $\{\mathbf{s}_o, \mathbf{u}_o\} = \{\mathbf{o}, \mathbf{f}_e\}$. 
No contact force is modeled for regrasping fingers, which break contact with the object.
The trajectory is divided into $\bm{\tau} = \{\bm{\tau}_c, \bm{\tau}_r, \bm{\tau}_o\}$.
We then write our trajectory optimization problem as
\vspace{-.2cm}
\allowdisplaybreaks 
\begin{align}
\min_{\substack{\mathbf{s}_1,\dots,\mathbf{s}_H; \\ \mathbf{u}_1,\dots,\mathbf{u}_H}} 
& \quad J_{g}(\bm{\tau}_o) + J_{smooth}(\bm{\tau}) + J_V(\bm{\tau}) \label{eq:trajopt} \\
\text{s.t.} \quad
& \mathbf{q}_{min} \leq \mathbf{q}_t \leq \mathbf{q}_{max} \notag \\
& \mathbf{u}_{min} \leq \mathbf{u}_t \leq \mathbf{u}_{max} \notag \\
& f_{contact}(\mathbf{s}_{c,t}, \mathbf{s}_{o,t}) = 0 \notag \\
& f_{kinematics}(\mathbf{s}_{c,t}, \mathbf{s}_{o,t}, \mathbf{s}_{c,t+1}, \mathbf{s}_{o,t+1}) = 0 \notag \\
& f_{balance}(\mathbf{s}_{c,t}, \mathbf{s}_{o,t}, \mathbf{s}_{c,t+1}, \mathbf{s}_{o,t+1}, \mathbf{u}_{c,t}, \mathbf{u}_{o,t}) = 0 \notag \\
& f_{friction}(\mathbf{s}_{c,t}, \mathbf{s}_{o,t}, \mathbf{u}_{c,t}) \leq 0 \notag \\
& f_{contact}(\mathbf{s}_{r,t}, \mathbf{s}_{o,t}) \leq -\delta, \quad t < H \notag \\
& f_{contact}(\mathbf{s}_{r,H}, \mathbf{s}_{o,H}) = 0 \notag \\
& \mathbf{q}_{r,t} + \Delta \mathbf{q}_{r,t} - \mathbf{q}_{r,t+1} = 0. \notag
\end{align}
All of these terms are unchanged from \cite{kumar2024diffusion} and \cite{yang2024multi} except for $J_V$. The cost term $J_g$ drives the object to the goal location, and $J_{smooth}$ promotes a smooth trajectory.
The last cost term $J_V$ uses our learned value function to guides the optimization, as detailed in \ref{sec:vf as cost}.
We use a $J_V$ of zero for baselines which don't use the value function.
$f_{contact} \leq - \delta$ guarantees that the regrasping fingers avoid contact with a threshold $\delta$ until the final time step.
The final constraint ensures that configurations and actions are consistent for the regrasping fingers that move in freespace.
The combination of the final three constraints results in regrasping behavior for the specified regrasping fingers.

We solve the trajectory optimization with Constrained Stein Variational Trajectory Optimization (CSVTO) \cite{power2024constrained}.  
We use this trajectory optimization formulation both to collect diverse data for training value functions, and also at deployment to solve each sub-task.
\vspace{-0.1 cm}
\subsection{Data Collection} \label{sec:data_collection}

To train the value function, we require data on sample task executions.
We collect datasets $D_c = \{(\tau, \zeta, \rho)\}$ for each contact mode $c$.
$\tau$ is a trajectory executing the sub-task with contact mode $c$, $\zeta$ contains sub-task dependent auxiliary inputs, and $\rho$ is a cost label, calculated using $J_g$ with modification.

To collect these data, we first sample an initial state composed of a reasonable random configuration for both the robot hand joints and the object as follows:
We choose a stable default configuration for the robot and object in which there is no penetration, and distribute the random initial states normally about this configuration.
We then step the simulator for a short time after sampling each initial state to resolve penetrations, and filter out infeasible samples. 
In a screwdriver turning task, this might occur if the screwdriver falls out of reach. 
Next, we run a precursory grasping trajectory optimization on this initial state, which solves \eqref{eq:trajopt} with all fingers regrasping.
Once again, we filter out samples where the robot has failed to stably grasp the object.
This initial state generation procedure allows us to achieve diverse starting grasps for the task with the correct contact mode.

From these initial states, we execute the contact sequence $C$ using the trajectory optimization.
To label our dataset, we use $\rho = \mathrm{min}(||\mathbf{o}_T -\mathbf{o}_g||_2^2, \rho_{\mathrm{max}})$, where $\rho_{\mathrm{max}}$ is a ceiling on $\rho$.
We use $\rho_{\mathrm{max}}$ because, beyond a certain discrepancy between final state and goal state, we deem the task failed and predicting the exact cost becomes irrelevant.

\vspace{-0.1 cm}
\subsection{Value Function Training}
An overview of the value function training process is shown in Fig. \ref{fig:method}a.
We learn a function $V_c(\mathbf{s}, t, \zeta) = \hat{\rho}$ for each sub-task with contact mode $c$, which estimates the total expected cost $\hat{\rho}$ of the full task given any state $\mathbf{s}$ of the sub-task, time index $t$ in the optimized trajectory, and auxiliary sub-task-dependent inputs $\zeta$.
We can then incorporate $V_c$ into $J$ as described in Section \ref{sec:vf as cost}, with the goal of solving the trajectory optimization with fewer iterations and better task performance. We use supervised learning to train $V_c$ on datasets $D_c$, with mean squared error loss function
\[
\mathcal{L} 
= \frac{1}{|D_c|}
  \sum_{(\tau, \zeta, \rho)\in D_c}
  \sum_{t=1}^H
  \Bigl(V_c(s_t, t, \zeta) - \rho\Bigr)^2,
\]
where $|D_c|$ is the total number of states in $D$ and $s_t$ is the state in $\tau$ at time step $t$.
The MLP networks are trained with mini-batch gradient descent and the Adam optimizer.

However, using a single neural network to represent $V_c$ at test time can lead to undesirable local minima in the trajectory optimization. 
As it can be challenging to collect a dataset which covers the full distribution of possible value function inputs, it is possible with a single neural network to get an erroneous prediction for an out-of-distribution input.
Therefore, it is helpful to use an ensemble of neural networks to model the epistemic uncertainty: 
the uncertainty due to a lack of data \cite{chua2018deep}.
The variance of the ensemble should be lower for inputs more similar to those in $D_c$ and provides a signal for how confident we can be in the model's prediction.
We use $V_c^i$ to refer to the $i$'th member of the ensemble.


\subsection{Value Function as a Trajectory Optimization Cost} \label{sec:vf as cost}

We use the learned value function at deployment time, as shown in Fig. \ref{fig:method}b.
In our trajectory optimization, each iteration involves calculating the gradient of the cost function and constraints with respect to the current trajectory and using those gradients to update said trajectory.
Our proposed method adds an additional cost component to the optimization.
To calculate this cost, we query $V_c$ at each optimization iteration, which gives $M$ predicted cost outputs (one for each ensemble member) for each of the $H$ states in the trajectory.
Then, we define cost $J_V(\tau)= \alpha \mu + \beta \sigma^2$, where
\vspace{-0.1 cm}
\[
\mu = \tfrac{1}{M}\sum_{t=1}^{H}\sum_{i=1}^{M} V_c^i(\mathbf{s}_t, t, \zeta),
\]
\vspace{-0.2 cm}
\[
\sigma^2 =\!\tfrac{1}{M} \sum_{t=1}^{H}  \sum_{i=1}^{M} \bigl(V_c^i(\mathbf{s}_t, t, \zeta)- \tfrac{1}{M}\sum_{j=1}^{M} V_c^j(\mathbf{s}_t, t, \zeta)\bigr)^2
\]
and \(\alpha, \beta > 0\) are scalar weights on the cost terms.

Minimizing the mean $\mu$ guides the optimization to select a trajectory that minimize the estimated future cost predicted by the value function ensemble for each state.
Minimizing the variance $\sigma$ drives the optimization toward states that are in distribution for $V_c$. 

\begin{figure}[t]
    \centering
    \includegraphics[width=1.0\linewidth]{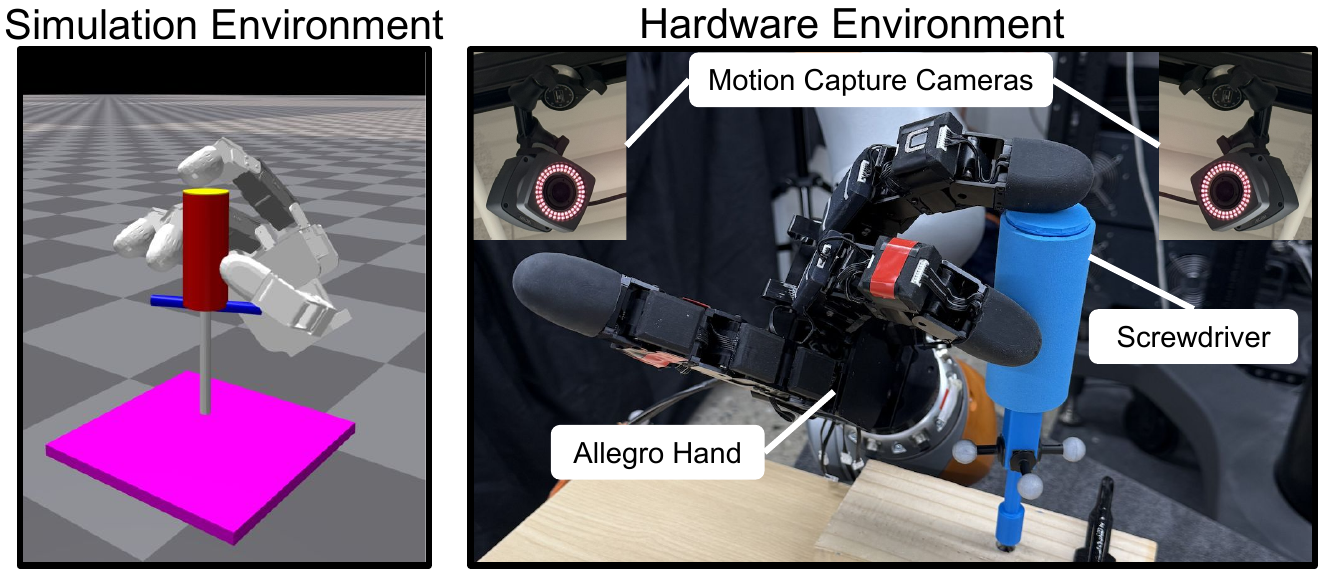}
    \caption{Simulation and hardware environment setups for our screwdriver turning task. In simulation, we have full access to the state variables. In hardware, the robot hand provides proprioceptive data, and we use observe the screwdriver orientation with motion capture cameras.}
    \label{fig:setup}
\end{figure}

\begin{figure*}[t]
    \centering
    \includegraphics[width=1.0\linewidth]{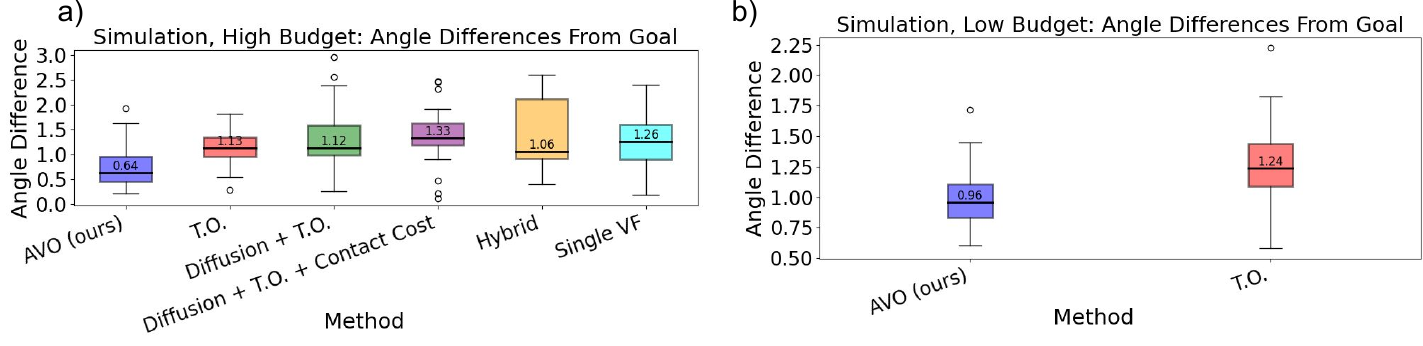}
    \vspace{-.7cm}
    \caption{Boxplots comparing quaternion angle differences between the final state and the goal state for each method during the High Budget (a) and Low Budget (b) simulation experiments. The median values for each method are labeled across 50 trials. }
    \label{fig:boxplot}
    \vspace{-.4cm}
\end{figure*}

\begin{figure}[t]
    \centering
    \includegraphics[width=1.0\linewidth]{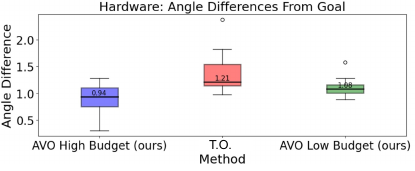}
    \vspace{-0.6 cm}
    \caption{Boxplot comparing quaternion angle differences between the final state and the goal state for each method during hardware evaluations. The median values for each method are labeled across 10 trials. }
    \label{fig:boxplot hardware}
\end{figure}

\vspace{-0.1 cm}
\section{Experiments and Results}

We evaluate our method on a screwdriver turning task, detailed in \ref{sec:sim_screw}, in simulation and hardware.
By evaluating against multiple baselines and ablations, we seek to show the utility of the learned value function in both (1) producing higher-performing trajectories and also (2) requiring less computation time to solve for successful trajectories.

We use a simulated Allegro hand \cite{allegro_hand_2024} in the Isaac Gym simulator \cite{makoviychuk2021isaac} for data collection and simulation experiments.
The screwdriver model is composed of two cylinders, and a spinning cap on the top so that it can turn freely while the index finger is placed on top (i.e a precision screwdriver).
\vspace{-0.2cm}
\subsection{Baselines}

We evaluate \methodname\ against five baselines and ablations in simulation. In the real world, we evaluate against the best-performing baseline from the simulation experiments.

Our first ``T.O'' baseline is our trajectory optimization method \cite{power2024constrained} with $J_V = 0$. This method does not make use of the collected data or value function in any way. Next, we include an ablation referred to as ``Single VF,'' which uses a single value function instead of an ensemble. We also implement an existing amortized trajectory optimization method, dubbed ``Diffusion+T.O.'' 
For this method, we adapt the work of Kumar et. al \cite{kumar2024diffusion}, which uses a diffusion model to generate full sub-task trajectories conditioned on the current state and contact mode.
These trajectories are then used as initializations to the same trajectory optimization problems used in our method.
From there, we roll out our receding horizon trajectory optimization as normal.
The dataset for the diffusion model consists of all of the ``successful'' trajectories from our value function dataset.
Filtering trajectories is important before training the diffusion as diffusion will sample trajectories similar to the training data, regardless of their quality.
To filter successful trajectories, we qualitatively select a threshold value $\rho_{successful}$.
As done in \cite{kumar2024diffusion}, we augment the diffusion dataset using a combination of task executions and CSVTO plans.
At each execution time step $t$ during data collection, we add a new trajectory to the diffusion dataset consisting of the $t$ executed states and the $H-t$ remaining planned states for the sub-task.

Additionally, we implement an extension of this method (``Diffusion+T.O.+Contact Cost'') with an additional cost component added to the trajectory optimization for the regrasping task.
This cost component is formulated as the norm between the planned fingertip positions and the fingertip positions at the last state of the diffusion-generated trajectories.
In \cite{kumar2024diffusion}, a similar cost is computed using the initial contact points of the hand on the object, as it is assumed the hand starts in a configuration conducive to turning.
However, this is not always true for our experiments as we randomly initialize $\mathrm{s}_0$.
Therefore, we use the diffusion to estimate where the hand should make contact with the object.
The cost term is influential as while the diffusion implicitly selects contact points, minimizing the other components of $J$ can lead the planned trajectory away from these contact points.
We use this alteration to investigate the utility of the contact points planned by the diffusion as compared to using the value function to implicitly optimize for these contact points.

Lastly, we include a hybrid method of diffusion and our value function ensemble (``Hybrid'').
Here, we initialize the trajectory optimization problem with the diffusion model, and then also use the $J_V$ cost term during optimization.

All baselines use the same computation budget and underlying trajectory optimization parameters as our method unless otherwise stated.
\vspace{-.1cm}
\subsection{Screwdriver Turning Task}\label{sec:sim_screw}

We evaluate \methodname\ on a screwdriver grasping and turning task.
Like many dexterous tasks in the real world, this task requires the robot to switch between several contact modes with the tool. 
Also like many dexterous tasks, it is a very precise and contact-reliant task, where the screwdriver can easily be dropped.
We begin the task in a stable configuration in which all three fingers are in sticking contact with the screwdriver (Fig. \ref{fig:overview}a).
Then, we execute a regrasping sub-task in which the thumb and middle fingers temporarily break contact with the object and then re-establish contact (Fig. \ref{fig:overview}b).
This mimics how a human first reorients their fingers in order to twist a screwdriver.
Then, we execute a ``turning'' sub-task, where all fingers are in contact and the robot seeks to turn the screwdriver ninety degrees clockwise (Fig. \ref{fig:overview}c).

We collected 10,000 samples for each sub-task's $D_c$.
We represent the screwdriver orientation $\mathbf{o}$ using $x$-$y$-$z$ Euler angles, denoted by $\phi$, $\theta$, and $\psi$.
We initialize $\mathbf{o}_0$ with $\mathbf{o}_\phi$ and $\mathbf{o}_\theta$ normally distributed about 0 with a standard deviation of 0.015 radians, and $\mathbf{o}_\psi$ uniformly distributed from $0$ to $2\pi$ radians.
The initial positions of the robot joints are also distributed normally about a default configuration as described in \ref{sec:data_collection}, with a standard deviation of 0.05. 

For this task, $\rho_{\mathrm{max}} = 5.0$, and we $M=16$ networks to model $V_c$.
For the turning sub-task (shown in Fig. \ref{fig:overview}a), we use $\zeta = \psi_0$, where $\psi_0$ refers to the value of $\psi$ at the first state of the turning sub-task.
When turning, $\mathbf{o}_g$ is dependent on $\mathbf{o}_0$ as we aim to turn for 90 degrees. 
Therefore, providing $\psi_0$ to the network avoids input aliasing issues that may occur without any goal specification. 
Neural network hyperparameters, cost function hyperparameters, $\alpha$, and $\beta$ are tuned via grid search.
For both sub-tasks, each MLP had a single hidden layer: the turning sub-task network used 32 neurons in its hidden layer, while the regrasp sub-task network used 24.

For the Diffusion+T.O. baseline, we select $\rho_{successful} = 1.0$, which results in a dataset consisting of 2,860 out of the 10,000 original full task trajectories.

\vspace{-.2cm}
\subsection{Experiments}
\vspace{-.1cm}

For evaluation, we execute the screwdriver turning task using each method, keeping the initial screwdriver and robot configurations constant.
For each experiment discussed below, we report the following values: (1) the median angle difference, defined as the magnitude of the 3D rotation that maps the final screwdriver frame onto the goal frame, and (2) the drop rate, which is the percent of trials during which the hand was unable to maintain all three points of contact with the screwdriver by the end of the task.

\noindent\textbf{Simulation Experiments:}

We use the same simulated environment for testing as we used for collecting training data.
We show our environment setups for both simulation and hardware experiments in Fig. \ref{fig:setup}.
For each simulation experiment, we report angle difference and drop rate over 50 random starting configurations.

\begin{table}[htbp]
\centering
\caption{Drop Rates for all Experiments}
\begin{tabular}{lcc}
\toprule
\textbf{Experiment} & \textbf{Method} & \textbf{Drop Rate $\downarrow$} \\
\midrule
\multirow{6}{*}{High Budget (simulation)} 
  & \methodname\ (ours) & \textbf{4.00}\% \\
  & T.O. & \textbf{4.00}\% \\
  & Diffusion+T.O. & 14.00\% \\
  & Diffusion+T.O.+Contact Cost & 32.00\% \\
  & Hybrid & 10.00\% \\
  & Single VF & 46.00\% \\
\midrule
\multirow{2}{*}{Low Budget (simulation)} 
  & \methodname\ (ours) & \textbf{4.00}\% \\
  & T.O. & 8.00\% \\
\midrule
\multirow{3}{*}{Hardware} 
  & \methodname\ High Budget (ours) & \textbf{0.00}\% \\
  & \methodname\ Low Budget (ours) & 10.00\% \\
  & T.O. & 30.00\% \\
\bottomrule
\end{tabular}
\label{tab:drop_rate}
\end{table}

Our first experiment, ``High Budget'' (Fig. \ref{fig:boxplot}a), provides ample computation budget for all methods. 
For the regrasp sub-task, we use 80 iterations for the first step to warm up the optimizer, and 25 for the remaining steps.
Fur the turning sub-task, we use 100 iterations for the first step and 25 for the rest.
Because all methods are able to converge to a solution within the budget provided, this experiment investigates the difference in solution quality across methods.

Our second experiment, ``Low Budget'' (Fig. \ref{fig:boxplot}b), reduces the optimization iterations by about 50\%.
For the regrasp sub-task, we use 40 iterations for the first step and 12 for the remaining steps.
Fur the turning sub-task, we use 50 iterations for the first step and 12 for the rest.
This experiment tests the amortization capabilities of \methodname, and whether \methodname\ can enable us to run trajectory optimization at a higher frequency than in the High Budget experiments. 

\noindent\textbf{Hardware Experiments:}

To replicate our simulation environment in the real world, we use the Allegro hand (Wonik Robotics) mounted to a static base, and we manufacture a screwdriver with the same size, shape, and spinning top cap as the simulated screwdriver.
During trials, the screwdriver turns a screw which sits loosely in a wooden board and therefore does not experience reaction forces from turning.
Proprioception from the hands provides the robot state, and we use a motion capture system (Vero, Vicon Motion Systems) to detect the screwdriver state.

Similarly to the simulation experiments, we use the same initial robot configuration for each trial across all methods. 
Because we cannot easily replicate the exact starting screwdriver configuration in real life, we attempt to insert the screwdriver in a perfectly upright position every time.

We selected the T.O. method as our baseline for the hardware experiment, because the simulation results showed T.O. to have the lowest drop rate of the baselines, and also among the lowest median angle differences.
We also test \methodname\ with both computation budgets, to show how the amortization results will transfer to real life. 
Results for 10 trials are reported in Fig. \ref{fig:boxplot hardware}.
\vspace{-.1cm}
\section{Discussion}
\vspace{-.1cm}

The results in Fig. \ref{fig:boxplot}a show that with a large computational budget, \methodname\ improved the quality of the optimized trajectories.
\methodname\ got 42.86\% closer to the goal (in terms of quaternion angle difference) compared to the next best method, which was Diffusion+T.O.
It also tied for the lowest drop rate of 4\%, along with T.O.
Although the large budget allows T.O. and other baselines to find a solution with a low $J$, executed trajectories can have lower performance than what the cost along would indicate.
This mostly results from the modeling errors of the optimization, e.g., discretization errors and quasi-static simplifications.
In contrast, our value function learns directly from simulated trajectories, guiding the optimizer to output trajectories that are more likely to have good performance when executed. 

Notably, \methodname\ was also able to outperform Diffusion+T.O. and Diffusion+T.O.+Contact Cost.
This indicates a benefit of learning from both low and high performing trajectories, unlike the diffusion method which is trained only on high performing trajectories.
Incorporating lower performing trajectories allows us to expand the dataset used to train our value function and provides information on gradients that can push lower performing trajectories to perform better.

Fig. \ref{fig:boxplot}b shows that the value functions were successful in amortizing the optimization process.
For the High Budget experiment, the T.O. baseline took 4.74 seconds on average per action, whereas \methodname\ took 5.07 seconds (6.47\% longer).
For the Low Budget experiment, the T.O. baseline took 2.38 seconds per action on average, and \methodname\ took 2.53 seconds (5.87\% longer).
This indicates 50.08\% less computation time for \methodname\ compared to the High Budget experiment, and 49.82\% less time for the T.O. method.
\methodname\ had a 22.58\% decrease in median angle difference from the goal compared to the T.O. method with a low budget, and critically a 50\% decrease in drop rate (Table \ref{tab:drop_rate}).
Furthermore, \methodname\ with a low budget outperformed every baseline with a high budget, with a 14.29\% improvement in angle difference over the best baseline, and an equal drop rate (4\%) to the best baseline.

Notably, both of the diffusion-based baselines underperformed, even though \cite{kumar2024diffusion} had success with these methods.
We believe this is primarily due to the diffusion model training on less data than \methodname, and also the state distribution for our task being larger than what was considered in \cite{kumar2024diffusion}.
For our tasks, we randomly sample the initial finger configurations, whereas \cite{kumar2024diffusion} had fixed initial finger configurations.
This wider range of initial states should require a larger training dataset, and it is possible that the training dataset was insufficient for the diffusion model to generalize to this wider distribution of inputs. 
Additionally, the diffusion initializations are quite aggressive when compared to \methodname\ and the T.O. baseline, and tend to heavily violate the constraints.
We believe this causes the Diffusion+T.O. method to be more failure-prone, as the optimizer is not always able to recover from these large constraint violations.
Although we hypothesized that the Diffusion+T.O.+Contact Cost baseline could guide the optimizer to choose better contact points, this method performed worse in both angle difference and drop rate compared to Diffusion+T.O., suggesting that the diffused contact points were either ineffective or destabilizing.


In our hardware experiments, we found that \methodname\ maintained an angle difference improvement over T.O.
As shown in Fig. \ref{fig:boxplot hardware}, the percent decrease with \methodname\ given the larger computational budget was 22.31\%, and 10.74\% with the smaller computational budget.
Furthermore, we found \methodname\ to be less failure-prone than the baselines---while T.O. dropped the screwdriver three out of ten times, \methodname\ (combining the high budget and low budget trials) dropped it only once out of twenty trials.
This suggests that \methodname\ is able to pick trajectories that are more stable and robust to failure by selecting better contact points when regrasping and by planning lower-cost trajectories when turning.†
The performance difference between \methodname\ with the high budget and low budget also remained small (the higher budget had 12.96\% lower angle difference), which is further evidence that \methodname\ is able to make a meaningful difference to the optimized trajectories in a short time. 


Although we only present results on a single task and contact sequence in this work, the design of \methodname\ can be applied to tasks with arbitrary contact sequences as well as robot geometry. Value function training data can be collected in simulation given an existing robot trajectory optimization pipeline, providing potential for our method to improve trajectory optimization among broad applications in dexterous manipulation. Although our task has a single reward signal based on the final state, value functions in other contexts have been used for different manipulation tasks with different formulations and frequencies of reward signals.
Furthermore, future work can consider appending additional sub-tasks with additional contact modes, or repeating sub-tasks in a loop. However, more evaluation is needed to verify the method's efficacy in these scenarios.

\vspace{-.1cm}
\section{Conclusion}
\vspace{-.1cm}

In this work, we present a method for amortized trajectory optimization in dexterous manipulation tasks with multiple contact modes. Our approach is successful on two fronts: it improves performance by bridging independently optimized sub-tasks, and lowers the computational budget needed to achieve a successful solution.
Our method consistently outperformed multiple baselines in both simulation and hardware experiments on a screwdriver turning task achieving superior results in both settings with just half of the computation budget.  
Future work will extend this approach to tasks with longer, more complex contact sequences.  We also plan to investigate coupling our method with contact-implicit trajectory optimization to potentially solve tasks without predefined contact sequences.

\bibliographystyle{ieeetr}
\bibliography{references.bib}

\end{document}